\documentclass[journal,twoside]{IEEEtran}

\usepackage{color}
\usepackage{amssymb}
\usepackage{rotating}
\usepackage[T1]{fontenc}
\usepackage{lscape}
\usepackage{dblfloatfix}
\usepackage{xcolor}
\usepackage{subcaption}
\usepackage[ruled,vlined]{algorithm2e}
\usepackage[autostyle]{csquotes}
\usepackage{tikz}
\usepackage{acro}
\usepackage{siunitx}
\usepackage{enumitem}
\usepackage{multirow}

\usepackage{hyperref}

\usepackage{tikz}
\usepackage{float}
\usepackage{algorithmic}
\usepackage{graphicx}
\usepackage{textcomp}
\usepackage{xcolor}
\usepackage{url}

\usepackage[autostyle]{csquotes}
\usepackage{acro}

\usepackage{multirow}

\usepackage{framed} 
\usepackage{nomencl} 
\usepackage[acronym,nomain,nonumberlist]{glossaries}
\usepackage[colorinlistoftodos,prependcaption]{todonotes}

\usepackage{comment}
\def\BibTeX{{\rm B\kern-.05em{\sc i\kern-.025em b}\kern-.08em
    T\kern-.1667em\lower.7ex\hbox{E}\kern-.125emX}}

\ifCLASSINFOpdf
\else
 
\fi

\usepackage[a4paper, total={170mm,257mm}]{geometry}

\begin{document}

%


\title{Unveiling the Invisible: Enhanced Detection and Analysis of Deteriorated Areas in Solar PV Modules Using Unsupervised Sensing Algorithms and 3D Augmented Reality}


\author{\IEEEauthorblockN{Adel Oulefki\IEEEauthorrefmark{1},
Yassine Himeur\IEEEauthorrefmark{2}, 
Thaweesak Trongtiraku\IEEEauthorrefmark{3}, 
Kahina Amara\IEEEauthorrefmark{1},
Sos Agaian\IEEEauthorrefmark{4},
Samir Benbelkacem\IEEEauthorrefmark{1},
Mohamed Amine Guerroudji\IEEEauthorrefmark{1},
Mohamed Zemmouri\IEEEauthorrefmark{5},
Sahla Ferhat\IEEEauthorrefmark{1},
Nadia Zenati\IEEEauthorrefmark{1},
Shadi Atalla\IEEEauthorrefmark{2},
Wathiq Mansoor\IEEEauthorrefmark{2}
}\\
\IEEEauthorblockA{\IEEEauthorrefmark{1}
Laboratory of LI3C, University of Biskra, Biskra, Algeria}\\
\IEEEauthorblockA{\IEEEauthorrefmark{2}College of Engineering and Information Technology, University of Dubai, Dubai, UAE}\\
\IEEEauthorblockA{\IEEEauthorrefmark{3}Faculty of Industrial Education Rajamangala, University of Technology Phra Nakhon, Vachira Phayaban Dusit Bangkok 10300, Bangkok, 10300, Bangkok, Thailand}\\
\IEEEauthorblockA{\IEEEauthorrefmark{4}Dept. of Computer Science, College of Staten Island, 2800 Victory Blvd Staten Island, New York, 10314, New York, USA}\\
\IEEEauthorblockA{\IEEEauthorrefmark{5}Université Kasdi Merbah, Ouargla, Ouargla, 30000, Ouargla, Algeria}\\
}


\maketitle

\begin{abstract}
Solar Photovoltaic (PV) is increasingly being used to address the global concern of energy security. However, hot spot and snail trails in PV modules caused mostly by crakes reduce their efficiency and power capacity. 
This article presents a groundbreaking methodology for automatically identifying and analyzing anomalies like hot spots and snail trails in Solar Photovoltaic (PV) modules, leveraging unsupervised sensing algorithms and 3D Augmented Reality (AR) visualization. By transforming the traditional methods of diagnosis and repair, our approach not only enhances efficiency but also substantially cuts down the cost of PV system maintenance. Validated through computer simulations and real-world image datasets, the proposed framework accurately identifies dirty regions, emphasizing the critical role of regular maintenance in optimizing the power capacity of solar PV modules. Our immediate objective is to leverage drone technology for real-time, automatic solar panel detection, significantly boosting the efficacy of PV maintenance. The proposed methodology could revolutionize solar PV maintenance, enabling swift, precise anomaly detection without human intervention. This could result in significant cost savings, heightened energy production, and improved overall performance of solar PV systems. Moreover, the novel combination of unsupervised sensing algorithms with 3D AR visualization heralds new opportunities for further research and development in solar PV maintenance.
\end{abstract}


\IEEEpeerreviewmaketitle

\section{Introduction}\label{sec1}
The generation of energy, particularly from fossil fuels, leads to the release of greenhouse gases such as carbon dioxide (CO2) into the atmosphere \cite{copiaco2023innovative,himeur2023ai}. These emissions contribute to global warming and climate change. By saving energy, we can decrease the demand for fossil fuel-based energy production, leading to lower greenhouse gas emissions and a reduced impact on the environment \cite{alsalemi2022innovative,elnour2022performance}.
Renewable energy, especially solar power, has emerged as a prominent solution in addressing global concerns related to climate change, unpredictable weather patterns, and the finite nature of fossil fuel resources \cite{himeur2022ai,khan2020investigating}. This has led to a significant increase in the deployment of photovoltaic (PV) power stations on a global scale \cite{benbelkacem2013augmented}. The growing recognition of the environmental and economic benefits of solar energy has fueled this expansion.
Solar power offers numerous advantages over traditional energy sources. Firstly, it is a clean and sustainable energy option, emitting minimal greenhouse gases during electricity generation. By harnessing sunlight, solar panels convert this abundant resource into usable electricity, reducing reliance on non-renewable fossil fuels and mitigating the environmental impact associated with their extraction and combustion \cite{zhang2023evaluating}.
Additionally, solar power installations have become more economically viable in recent years. Advancements in PV technology, coupled with declining costs, have made solar energy increasingly affordable and competitive with conventional energy sources. As a result, governments, businesses, and homeowners are investing in solar power systems to reduce energy costs, achieve energy independence, and contribute to a greener future \cite{rashid2022future}.
The global trend towards solar power adoption has been driven by various factors. Increasing public awareness of climate change and environmental conservation has created a growing demand for sustainable energy solutions. Moreover, government policies and incentives, such as feed-in tariffs and tax credits, have encouraged the adoption of solar power systems by providing financial support and favorable regulatory frameworks \cite{rashid2022future}.
As a consequence of these developments, the installation of PV power stations has witnessed a remarkable surge across the globe. Solar farms and large-scale PV installations are being constructed to meet the growing demand for clean energy. This expansion not only facilitates the transition towards a low-carbon economy but also stimulates job creation and local economic growth \cite{xia2022mapping}.

In addition to their widespread use, PV modules hold a pivotal role in determining the overall efficiency of a solar power station. However, over time, these modules are prone to a range of defects that can significantly impact their power output efficiency. Ideally, all the PV cells within a string should possess similar electrical characteristics and operate at the maximum power point (MPP) current, thereby optimizing their individual performance \cite{haque2019fault,himeur2022next}.
Unfortunately, variations in the electrical characteristics of the PV cells can occur, resulting in a mismatch in the string's current. This mismatch prevents the entire string from operating at each cell's MPP, leading to a suboptimal performance \cite{dwivedi2020advanced}.
One common factor contributing to the reduced efficiency of PV modules is the phenomenon known as Potential-Induced Degradation (PID). PID occurs when the PV module is exposed to high voltage differentials between its conductive elements and the ground. This can lead to leakage currents and subsequent degradation of the module's electrical properties, resulting in reduced power output \cite{ibne2023impact}.
Other factors that can affect the performance of PV modules include module soiling, shading, degradation of the anti-reflective coating, and hotspots caused by localized heating. Each of these issues can impact the overall efficiency of the solar power station and lead to suboptimal energy generation \cite{arosh2023composite}.

On the other hand, there are several factors that can cause variations in the electrical characteristics of PV modules, such as partial shading \cite{PALMER2019989} and short-circuited bypass diodes. When a low-current PV cell is present in a string of high short-circuit current PV cells, the forward bias across all the cells can reverse bias the shaded cell. This, in turn, significantly increases the temperature of the affected cell, leading to a phenomenon known as hot spotting. Hot spotting can not only damage the cell but also diminish the overall power output of the solar panel \cite{dhimish201970}. 
Therefore, conducting regular inspections of PV modules is crucial to ensure optimal output efficiency. Various methods are employed for PV module inspection, including manual inspection, laser detection for pinpointing potential issues with greater accuracy, satellite observations for obtaining a comprehensive view of the entire setup, infrared thermography for detecting anomalies in heat distribution, and electroluminescence imaging for identifying cracks or other defects that may not be visible to the naked eye \cite{aghaei2022review}. These inspection techniques help identify and address any performance issues or potential risks promptly, allowing for timely maintenance and maximizing the overall efficiency and lifespan of the PV system.

Manual inspection is a laborious process, while techniques such as laser detection and electroluminescence imaging are not suitable for large-scale PV power stations. Infrared thermography is widely adopted for inspecting large PV systems due to its ease of use. However, even with infrared thermography, inspecting a large PV system can be time-consuming as each module needs to be individually inspected \cite{trongtirakul2022unsupervised}. 
Recently, artificial intelligence has been extensively utilized for anomaly and fault detection \cite{himeur2021artificial,himeur2020novel}. In the field of Solar PV Modules Build-up, detecting abnormalities using AI, drones, virtual reality, and other technologies has emerged as a prominent research area.
For instance, drone-based infrared thermography has gained considerable attention as a promising approach to streamline the inspection process of PV systems. By utilizing drones equipped with infrared cameras, it becomes possible to efficiently capture thermal data of the entire PV system from an aerial perspective \cite{fahimipirehgalin2021automatic}. This technology offers several advantages, such as improved accessibility to hard-to-reach areas and the ability to cover large areas quickly.
However, despite its potential, many existing approaches in drone-based infrared thermography still face certain limitations. One major drawback is the reliance on manual drone control, which can be physically demanding and time-consuming. Piloting the drone manually requires skilled operators who must navigate the drone precisely to capture thermal images of all PV modules. This process can be challenging, particularly for large-scale installations that encompass numerous modules \cite{henry2020automatic}.
Moreover, a significant issue with existing approaches is the lack of precise information about the location of defective panels. While thermal images obtained by drones can identify areas with abnormal temperatures, they often fail to provide accurate localization of the specific panels that require maintenance. As a result, the subsequent identification and repair of defective panels become more complex and time-consuming, leading to additional delays in the maintenance process \cite{masita202275mw}.


To overcome the aforementioned issues, the design, installation, and end-users of photovoltaic (PV) systems can all reap the benefits of augmented reality (AR) visualization. AR offers the potential for a highly realistic and immersive experience of the PV system, which can facilitate adjustments based on real-time usage and enhance the overall efficiency of the system. This, in turn, assists designers and installers in conducting more accurate analyses and making improvements to the system's design. Moreover, AR has the potential to enhance users' understanding of equipment maintenance and operation, thereby improving their overall experience and proficiency in utilizing the PV system. Additionally, visualization tools improve the localization of deteriorated areas in Solar PV systems by providing enhanced imaging, real-time monitoring, data analysis, augmented reality, 3D modeling, and historical data comparison \cite{al2022interactive}.
Furthermore, the implementation of AR can greatly support PV system installation and maintenance processes \cite{AR-main}. By providing installers with a visual representation of the system in action, AR simplifies the installation and positioning of panels and other components. Additionally, AR technology aids in the identification and diagnosis of any potential issues or maintenance requirements, reducing downtime and optimizing system performance. Overall, the integration of AR in PV systems brings numerous advantages, enhancing the efficiency, effectiveness, and user experience throughout the system's lifecycle \cite{al2021solar,al2020energy}.

Specifically, AR can serve as a powerful tool for promoting the adoption of renewable energy. By providing a delightful and immersive experience \cite{benbelkacem2022covi3d}, AR has the potential to inspire individuals to embrace sustainable energy practices and increase their understanding of PV technology \cite{AR-PV-Main}. The utilization of AR in visualizing PV systems offers numerous advantages as it delivers vital information regarding their construction, installation, and maintenance. This, in turn, can contribute to a broader acceptance of renewable energy sources.
The combination of augmented reality and infrared thermography presents a comprehensive solution for efficiently monitoring and diagnosing faults in PV modules, thereby enhancing their overall performance and lifespan. By integrating these technologies, the study introduces a novel method for detecting and localizing faults in PV modules utilizing infrared thermography. The proposed method encompasses the following key contributions:

\begin{itemize}

\item Evaluating the condition of PV modules to determine if they are functioning normally or if there are any defects present.

\item Creating a solution for detecting and locating faults in PV modules by employing improved segmentation techniques and visualizing 3D thermal images sourced from the Cali-Thermal Solar Panels Database.

\item Introducing a novel approach for enhancing and segmenting PV images to effectively handle irregularities or anomalies.

\item Introducing an advanced system based on Augmented Reality for 3D visualization and localization, which forms an integral part of the proposed method.

\end{itemize}

The rest of this paper is organized as follows. In Section \ref{Literature Review}, we delve into a comprehensive literature review, underscoring the necessity for novel methodologies in this domain. Section \ref{Methodology} elucidates our proposed method of abnormality analysis, complete with an insight into data collection and pre-processing strategies. Subsequently, Section \ref{Results} showcases the evaluation outcomes and performance comparison of our approach. Ultimately, Section \ref{Conclusion} encapsulates the significant findings of the study.

\section{Literature Review}
\label{Literature Review}

Various techniques have been proposed for damage detection on solar panels. In this section, we provide an overview of some existing techniques and highlight their key characteristics.
For instance, Alsafasfeh et al. \cite{alsafasfeh2018unsupervised} proposed a technique that combines thermal and visual data imagery to detect various faults. They employed the Canny edge detector, Gaussian filter, and histogram equalization along with seed pixels to identify faults. This technique offers real-time monitoring capabilities for PV system operations and can detect various types of faults. However, it does not specifically address dust-related issues.
Similarly, Shihavuddin et al. \cite{shihavuddin2021image} also developed a technique that utilizes thermal and visual data imagery for fault detection. They employed a single trained model capable of detecting different types of damage and provided a new dataset comprising four specific image sets. While this technique shows promise in detecting various types of damage, the use of a single model may reduce sensitivity to different types of damage.
Moving forward, Zyout et al. \cite{zyout2020detection} proposed a technique for surface defect detection using online visual images. They employed AlexNet and CNN convolutional neural networks to classify the images. This technique introduces an innovative concept but relies on manual feature extraction during the detection stage. Furthermore, relying on online data collection may limit the capacity of the classification model.

Furthermore, Henry et al. \cite{henry2020automatic} presented a technique that leverages thermal and visual data imagery to detect deteriorated PV panels. They employed color-based segmentation followed by contour detection to identify faults. The approach was extensively evaluated using a large real-world dataset. However, it should be noted that the determination of the root cause of the detected fault still requires manual intervention.
In a different approach, Abuqaaud et al. \cite{abuqaaud2020novel} proposed a technique for dust and soil detection using RGB cameras. They employed the Gray Level Co-occurrence Matrix (GLCM) method for image classification. The technique is relatively straightforward to implement, but it does not account for other classes of anomalies such as shadow areas, broken panels, or wet panels.
Additionally, Pierdicca et al. \cite{pierdicca2020automatic} presented a technique for anomaly cell detection utilizing a thermal infrared sensor. They employed the Mask R-CNN architecture for image classification. The technique includes a publicly accessible dataset and has been compared to recent works employing deep neural networks. However, there is room for improvement by incorporating real-time electrical data analysis from operating photovoltaic modules using a monitoring infrastructure.


Segmenting deteriorating areas of a PV system has the advantage of accurately identifying and diagnosing any problems. With the help of AR technology, maintenance workers can quickly locate and evaluate damaged regions in real time. This means that inspections take less time and repairs can be done faster, resulting in less downtime. AR can also be useful during maintenance and repairs by providing workers with step-by-step instructions on how to fix issues. By superimposing repair instructions and schematics onto the AR display, maintenance workers can reduce the likelihood of making mistakes and improve the quality of repairs \cite{benbelkacem2013augmented}.

Even though, AR technology can enhance safety during maintenance procedures by visualizing potential hazards and safety issues, enabling maintenance staff to take necessary precautions and avoid accidents that may cause injury to persons or damage to property, including the PV system.
Regardless, despite the growing interest in using AR for various industrial maintenance tasks, there remains a noticeable research gap in exploring the potential of AR for enhancing the maintenance of solar panel PV systems.
In addition, AR can decrease the cost of PV system maintenance and repair by facilitating more efficient diagnosis and repair procedures and reducing labor costs and downtime. Moreover, AR provides real-time information on the PV system's condition, preventing minor issues from escalating into larger problems, and ultimately extending the system's lifespan and decreasing the need for costly repairs. In summary, AR visualization of damaged PV system components offers several advantages, including increased safety, cost savings, and more effective maintenance and repair processes.

\section{Methodology}
\label{Methodology}

\begin{figure*}[h]
\includegraphics[width=0.9\textwidth] {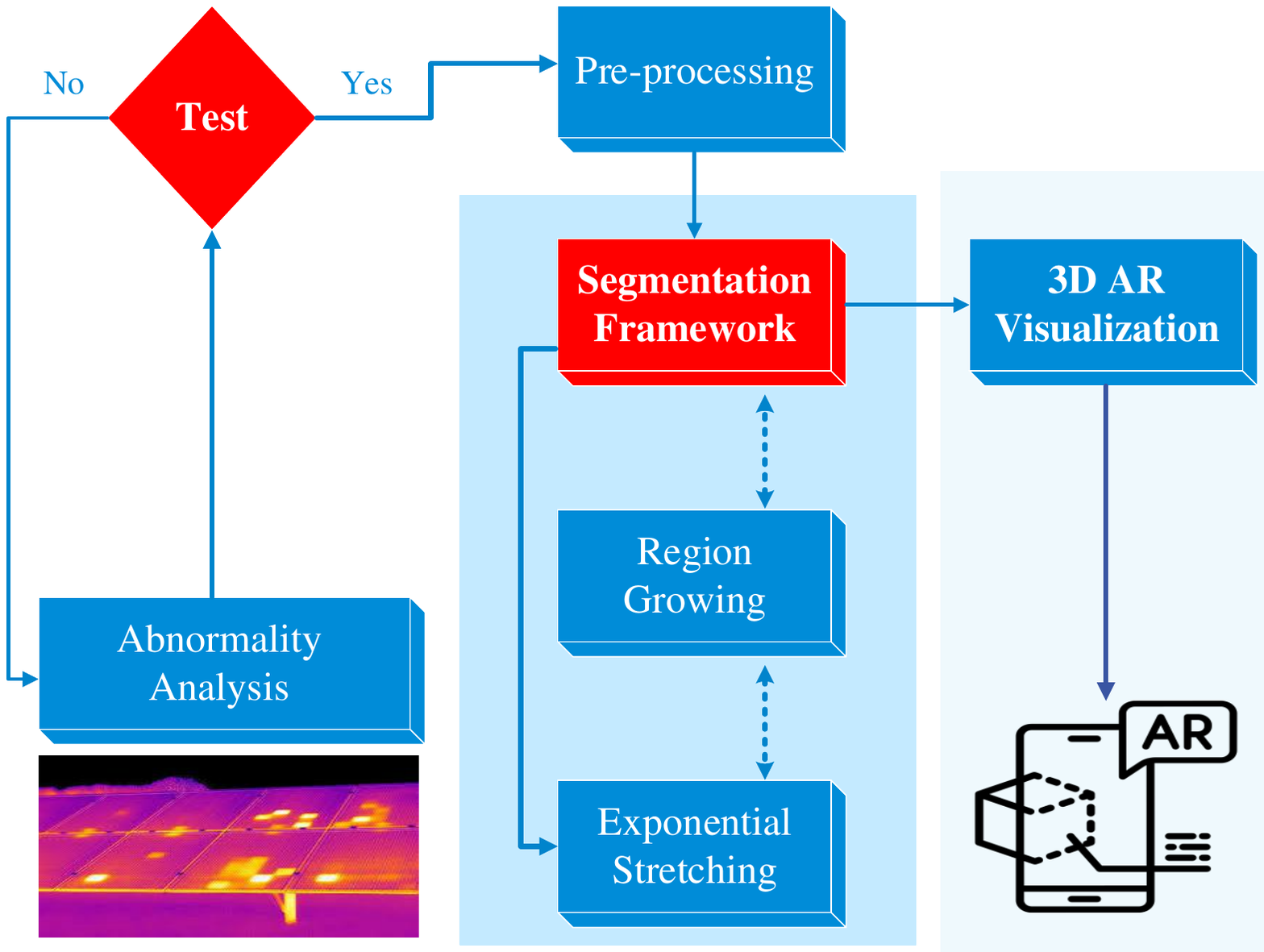}
\caption{ Block Diagram of the methodology used for analyzing and visualizing deteriorated area in PV module }
\label{fig:01}
\end{figure*}

This section introduces key methodologies for analyzing and visualizing abnormal data in a 3D environment. We cover the abnormality analysis, data collection, pre-processing, 3D broken area tracking using a segmentation framework with exponential stretching function and Region Growing-Based Segmentation, and 3D Augmented Reality visualization and localization. Each technique is essential for the accurate analysis and visualization of abnormality data, and we will delve into their details in the following subsections. Figure \ref{fig:01} illustrates the block diagram of the methodology used for analyzing and visualizing deteriorated areas in a PV (photovoltaic) module

\subsection{Abnormality analysis method} 

the Abnormality analysis method is an essential step for saving time before proceeding to segmentation. In our case, we rely on thermal image analysis, referring to the method developed in \cite{henry2020automatic} for detecting abnormal PV modules. This process determines PV module health based on criteria in equation (5). Using the thermal image, we determine the highest temperature value $T_{max}$, lowest temperature value $T_{min}$, and mean temperature value $T_{m}$. Subsequently, either the high-temperature threshold value $T_{h}$ or the low-temperature threshold value $T_{l}$ is calculated using Equations (2) and (3), respectively.

\begin{equation}
    T_{h} = T_{mean} + (T_{max} \times 0.2)
\end{equation}

\begin{equation}
    T_{l} = T_{mean} + (T_{min} \times 0.2)
\end{equation}

If the temperature value $T_{val}$ is less than $T_{max}$ and $T_{min}$, the count value $a_{c}$ is raised.

\begin{equation}
    f(x) = \left\{\begin{matrix}
a_{c}+ + & if (T_{val} > T_{max}) or (T_{val} < T_{min}) \\ 
a_{c} & otherwise
\end{matrix}\right.
\end{equation}

The module is classified as abnormal if $a_{c}$  is greater than 0.2\% of the module area value $S_{module}$.

\begin{equation}
    g(x) = \left\{\begin{matrix}
abnormal PV module & if (a_{c} > S_{module} \times 0.002) \\ 
normal PV module & otherwise
\end{matrix}\right.
\end{equation}

The equation presented above enables the identification of faulty PV modules in a large-scale PV power station. This allows us to focus the subsequent segmentation process specifically on the abnormal PV modules, saving time and resources by narrowing down the analysis to the relevant areas of interest.

\subsection{ Data collection }

\begin{figure*}[h]
\centering
\includegraphics[width=\linewidth]{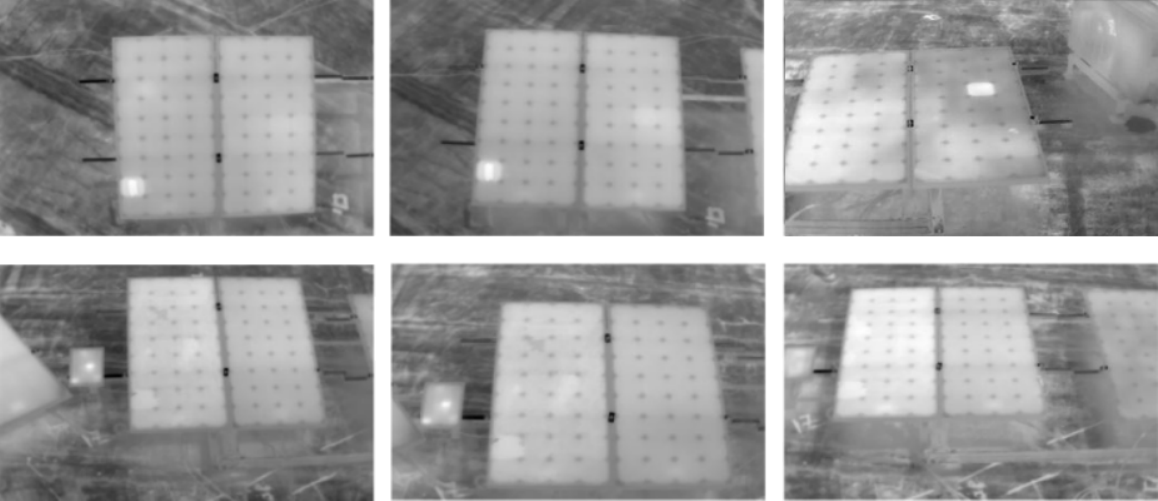}\\
(a)\\
\includegraphics[width=\linewidth]{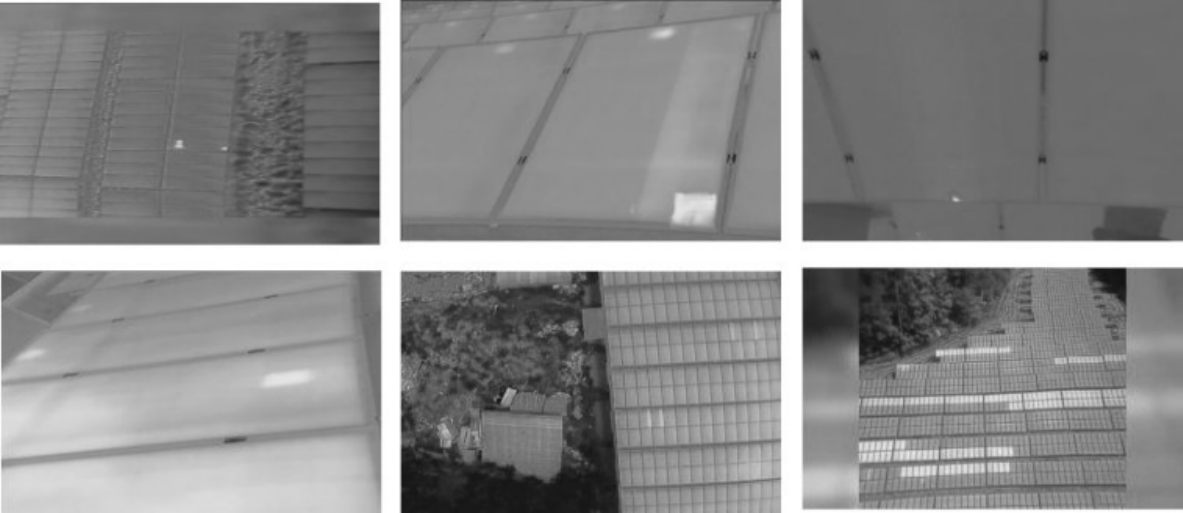}\\
(b)\\
\caption{Samples of abnormal and normal PV modules highlighting the distinct differences in visual appearance and potential defects on (a) Cali-Thermal Solar Panels and (b) Solar Panel Infrared
Image Database }
\label{fig:02}
\end{figure*}

Prior to the segmentation and visualization processes, data collection was conducted using the Cali-Thermal Solar Panels Image database \cite{alfaro2019dataset,gallardo2017aerial}. This comprehensive dataset includes a wide range of test images depicting various areas of solar panel deterioration, as illustrated in Figure \ref{fig:02} (a). The dataset encompasses diverse scenarios of deteriorated cases, encompassing different types of deterioration or sample PV panels. Indeed, Figure \ref{fig:02} serves also as a visual representation of the differences in visual appearance between abnormal and normal PV modules, providing a visual reference for identifying potential defects in this database.

The images serve as the foundation for the proposed segmentation and visualization techniques, allowing for accurate analysis and visualization of abnormal data in the context of solar panel deterioration.

On the other hand, we also picked the dataset solar panel infrared images v5 \cite{solar-panel-infrared-images-v5} for segmentation purposes. The dataset includes 934 images of solar panels, which are annotated in Tensorflow Object Detection format. Each image has been resized to a resolution of ($416x416$) pixels. This dataset is designed for computer vision projects related to solar panel inspection and defect detection \cite{SPII_DatasetName,vega2020solar}.  In Figure \ref{fig:02} (b), representative samples of the dataset are depicted, providing a visual representation of the underlying data.

\subsection{ Segmentation framework for tracking 3D broken areas}

The histogram stretching technique plays a crucial role in spatial domain pre-processing methods, which are essential for enhancing images, recognizing patterns, and performing binarization and segmentation. The linear stretching approach is widely employed for expanding luminance levels uniformly. However, its effectiveness is limited when the luminance levels are fully distributed. To overcome this limitation, non-linear techniques are utilized to compress some dynamic luminance levels while expanding others.

\subsection{Exponential Stretching Function}

In this paper, we propose an exponential stretching function to expand the bright region. The function can be described as follows:

\begin{equation}
    f(x) = x_{L-1} \cdot \left(1-e^{-\frac{x-x_{min}}{x_{max}-x_{min}}}\right)
\end{equation}

Where $x$ refers to a luminance level, $x_{(L-1)}$ denotes the total number of luminance levels within a permitted range, $x_{min}$ and $x_{max}$ represent the minimum and maximum luminance levels, respectively.

\subsection{Region Growing-Based Segmentation}

This technique utilizes the region-growing method to combine image pixels, in which the starting point is divided into multiple locations. The algorithm calculates the region of interest into multiple regions and identifies redundancies. Finally, redundant regions are displayed in different colors. Algorithm 1 can be expressed as:

\begin{algorithm}[t!]
\caption{Region Growing-Based Segmentation}
\textbf{Input:} Stretched image, $S_{(i,j)}$\
\\
\textbf{Output:} Binary image, $B_{(i,j)}$, and segmented regions, $C_{(i,j)}$\
\\
$p_{(i,j)}=1; S_{(i,j)}\geq \tau$ 
\\
\textbf{For} $a=1$ \textbf{to} $i\times j$ 
\\
\quad $R_{(a,k)}=p_a$\
\\
\quad \vline~\textbf{while} $\dfrac{1}{N} \sum_{k=1}^{N} R_{(a,k)} \leq \varepsilon$ 
\\
\quad\quad\quad $R_{(a,k)}$ is connected to its surrounding pixels, $k=1,2,\ldots,N$.\
\\
\quad \vline~
\textbf{end while}\
\\
\quad $R_a=\bigcup_{k=1}^{N} R_{(a,k)}$ 
\\
\textbf{end}\


$B_{(i,j)}=\begin{cases} 1, & R_a<\dfrac{1}{N}\sum_{k=1}^{N} R_{(a,k)} \ 0, \& \text{else} \end{cases}$\
\\
$R_{(i,j)}=\prod_{a=1}^{i\times j} R_a$ 
\\
$C_{(i,j)}\leftarrow f_m (R_{(i,j)})$ 
\end{algorithm}


\subsection{ 3D Augmented Reality visualization and localization approach}

In order to improve the visualization of PV systems and identify deteriorated areas, we have developed a 3D model of a Solar Photovoltaic panel. The model was created using a combination of SolidWorks \cite{SolidWorks} and Blender \cite{Blender} software. Our design process consisted of three main steps: firstly, we created a base for the panel, followed by adding solar cells, and finally, we included details such as mounting holes, rounded edges, fasteners, textures, and colors to achieve a more realistic appearance. This 3D model serves as a valuable tool for studying and analyzing the behavior of PV systems under different conditions, and it can aid in the identification and diagnosis of areas that may require maintenance or repair.

\begin{figure*}[h]
\centering
\includegraphics[width=0.9\textwidth] {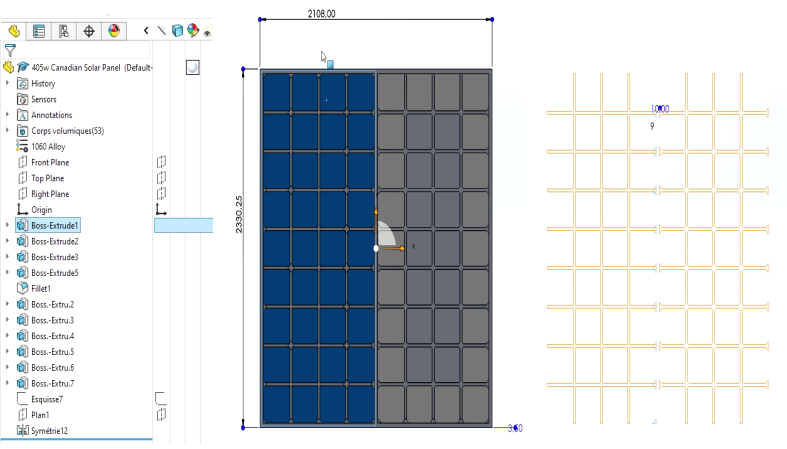}\\
(a)\\
\includegraphics[width=0.9\textwidth]{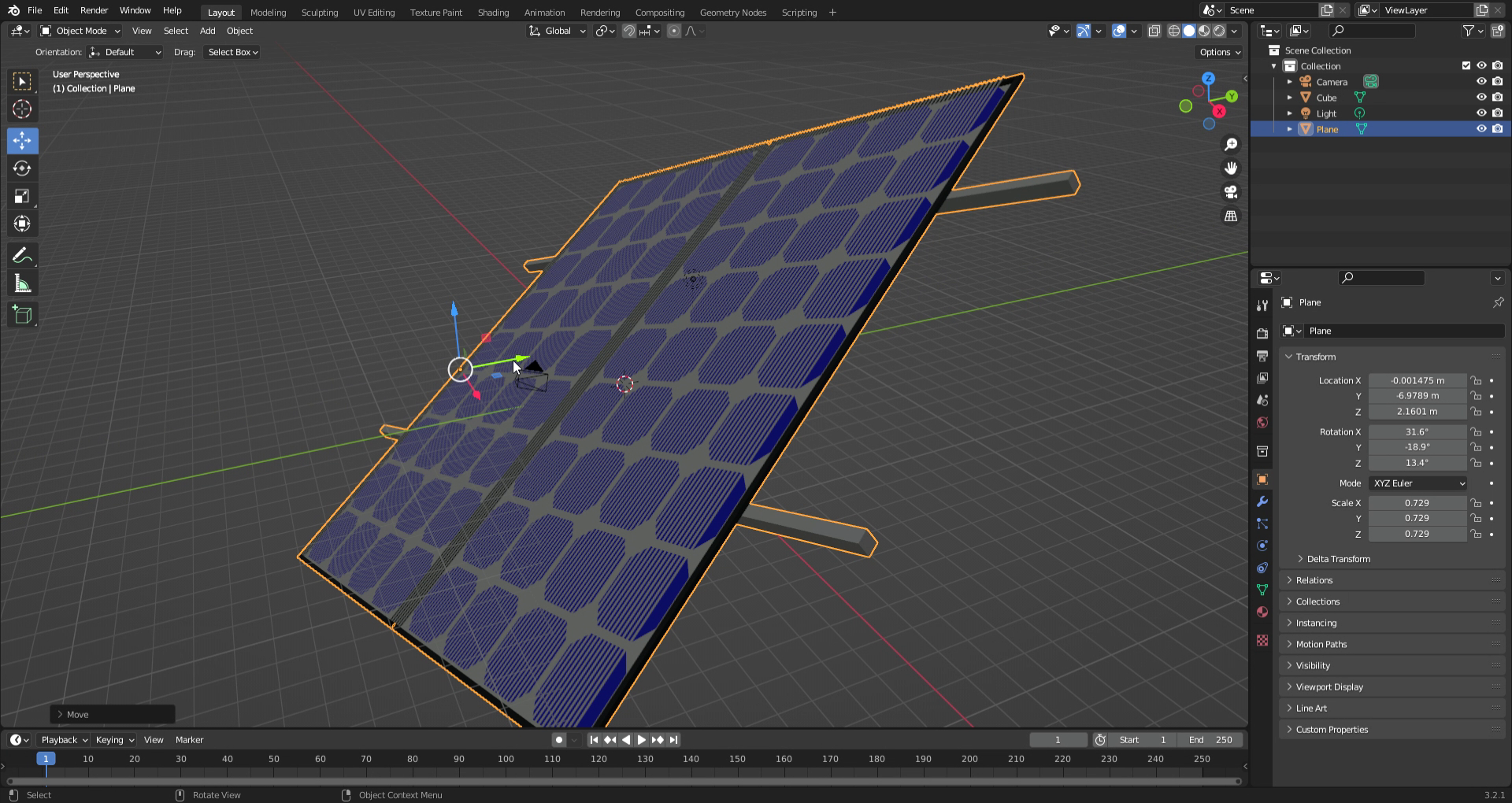}\\
(b)\\
\caption{ (a) Solidworks conception, (b) PV Blender conception}
\label{fig:031}
\end{figure*}

To begin with, we followed a straightforward three-step process in SolidWorks. Firstly, we created a new SolidWorks document. Secondly, we sketched the shape of the solar panel using 2D drawing tools such as lines, circles, and arcs. Finally, we applied features such as extrusions and cuts to convert the 2D sketch into a detailed and accurate 3D model of the solar panel. This process allowed us to easily and precisely manipulate the design and iterate on various options until we achieved the desired outcome. Overall, the use of SolidWorks significantly streamlined our design process, resulting in a highly efficient and effective design (see Figure \ref{fig:031}(a)).


The photorealistic appearance of our PV design was significantly enhanced through the use of Blender \cite{Blender}. By incorporating a variety of elements such as textures, colors, mounting holes, and softened edges, we were able to bring our concept to life. The resulting 3D model is highly detailed and visually stunning (see Figure \ref{fig:031}(b)). Additionally, Blender's versatile rendering features allowed us to experiment with different lighting and shading settings to achieve the best visualization performance.

Compared to using SolidWorks alone, Blender provided us with more flexibility and options for creating a more realistic and detailed PV design. Its ability to produce photorealistic textures and colors allowed us to better envision the final product. Furthermore, the incorporation of mounting holes and softened edges improved the design's usability and functionality. Overall, the combination of SolidWorks and Blender provided us with an efficient and effective approach to producing a high-quality PV design.


\subsubsection{\textbf{AR rendering}}

We utilized Vuforia SDK \cite{Vuforia} to enhance the realism and interactivity of our PV design visualization through augmented rendering. This cross-platform SDK provided robust tools for tracking and augmenting virtual objects in the real world. By utilizing markers as 3D features, we were able to precisely track and modify the design in real-time within the physical environment. These markers, specific forms, and images served as standards for scene localization and augmentation to ensure accurate tracking. With Vuforia's augmented virtual rendering, we achieved a highly realistic representation of the PV design in situ. This allowed us to better understand how the design would interact with its environment, a crucial consideration for evaluating its performance. Unity3D \cite{Unity} was used to create the AR environment, as it allows for full access to any item created and can import 3D models (.FBX) necessary for loading our PV 3D model. Our 3D reconstruction of the segmented deteriorated PV areas was also incorporated.


\section{Results} 
\label{Results}

To accurately detect and locate deteriorated cells within PV panels, it is essential to conduct an in-depth abnormality analysis. Before proceeding with the segmentation process, an evaluation of the abnormality analysis results must be performed to identify the specific panels/cells that require further investigation. Once the panels with deteriorated cells are identified, the proposed approach can be segmented and then evaluated using computer simulations to ensure its effectiveness.

Upon evaluation, the next step involves segmenting the PV panels to isolate the regions with deteriorated cells. This process is crucial in accurately identifying the damaged areas and preventing false positives. The segmentation results are then analyzed to obtain a comprehensive understanding of the extent of the damage.

To provide a more intuitive understanding of the damaged regions, AR visualization techniques can be employed. The AR visualization results enable users to visualize the damage in real-time, providing a clearer picture of the damage. In brief, in this section, we discuss the abnormality analysis results, evaluation of the proposed approach using computer simulations, segmentation results, and AR visualization results.

\subsection{Abnormality analysis results}

\begin{figure*}[h]
\centering
\includegraphics[width=12cm, height=7cm]{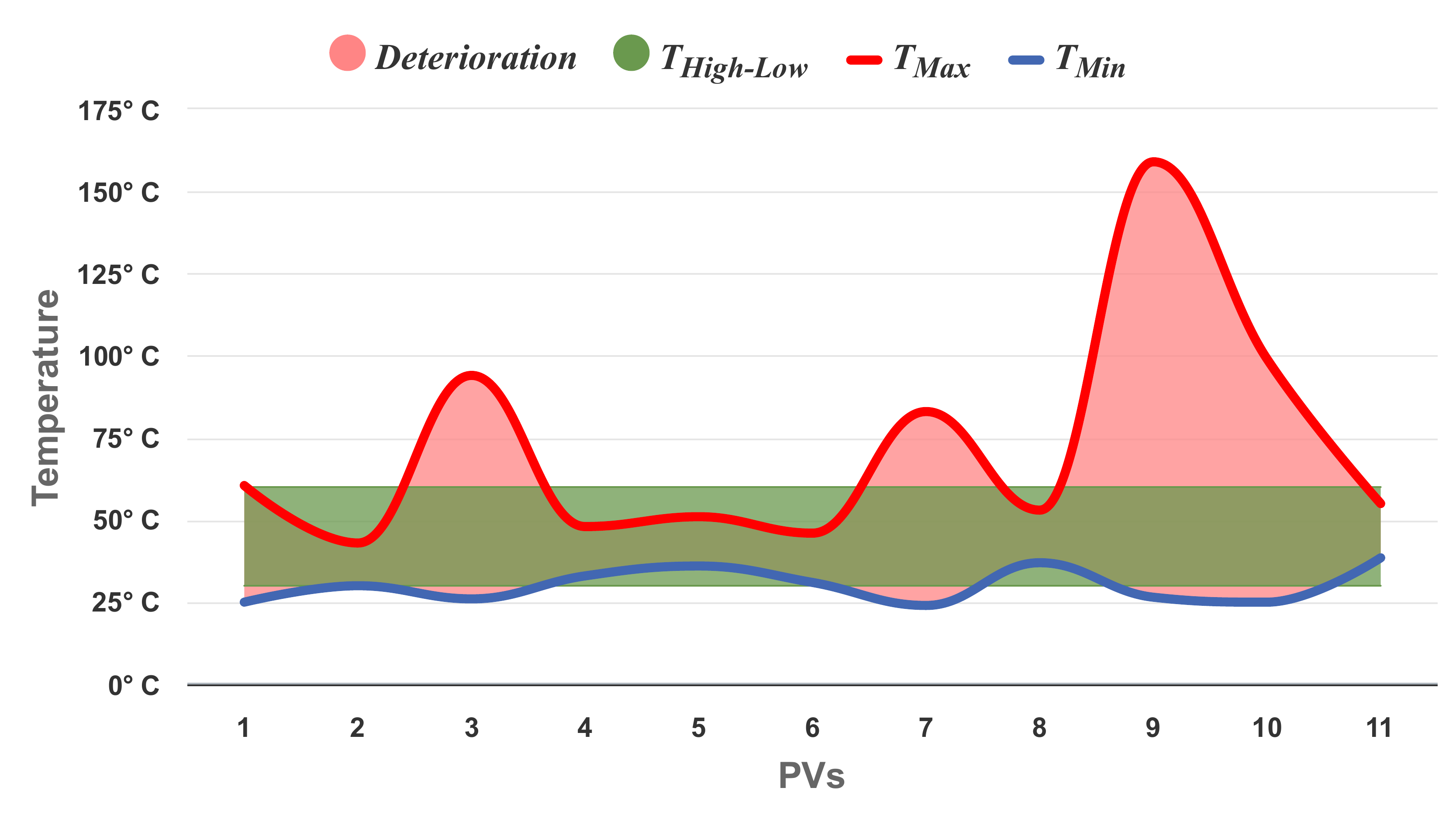}
\caption{ $T_{max}$ and $T_{min}$ values for 11 PV panels, indicating which panels have deteriorated cells. Segmentation is necessary to accurately delineate the affected region and optimize performance}
\label{Fig:055}
\end{figure*}

Figure \ref{Fig:055} illustrates the $T_{max}$ and $T_{min}$ values of 11 PV panels, providing information on their states. The panels with deteriorated cells can be identified by analyzing the obtained values, which is important for selecting the panel to which the segmentation approach will be applied. This can significantly reduce processing time. Panels 2, 4, 5, 6, 8, and 11 exhibit thermal values ($T_{val}$) within the range $T_{min}$ $<$ $T_{val}$$<$ $T_{max}$ ($25$$^{\circ}$C $<$ $T_{val}$$<$ $40$$^{\circ}$C), indicating that these panels are clean and do not require segmentation based on the approach proposed in subsection $A$. In contrast, panels 1, 3, 7, 9, and 10 display thermal values outside this range ($T_{val}$$<$ $T_{min}$ and $T_{val}$$>$ $T_{max}$), indicating deteriorated cells. These panels experience unfavorable operating conditions, such as increased heat stress or reduced efficiency, leading to higher and lower temperature values. The $T_{max}$ values for these panels are higher than those of the other PVs, suggesting that one or several cells have deteriorated. Similarly, the $T_{min}$ values for these panels are lower, indicating that they integrate cells with very low performances. The analysis of Figure \ref{Fig:055} emphasizes the importance of identifying the specific reasons for PV panel performance deterioration and implementing appropriate measures to mitigate the negative effects and optimize performance. Therefore, segmentation is necessary to accurately delineate the affected region

\subsection{Visual inspection results}

\begin{figure*}[h]
\centering
\includegraphics[width=0.9\textwidth]{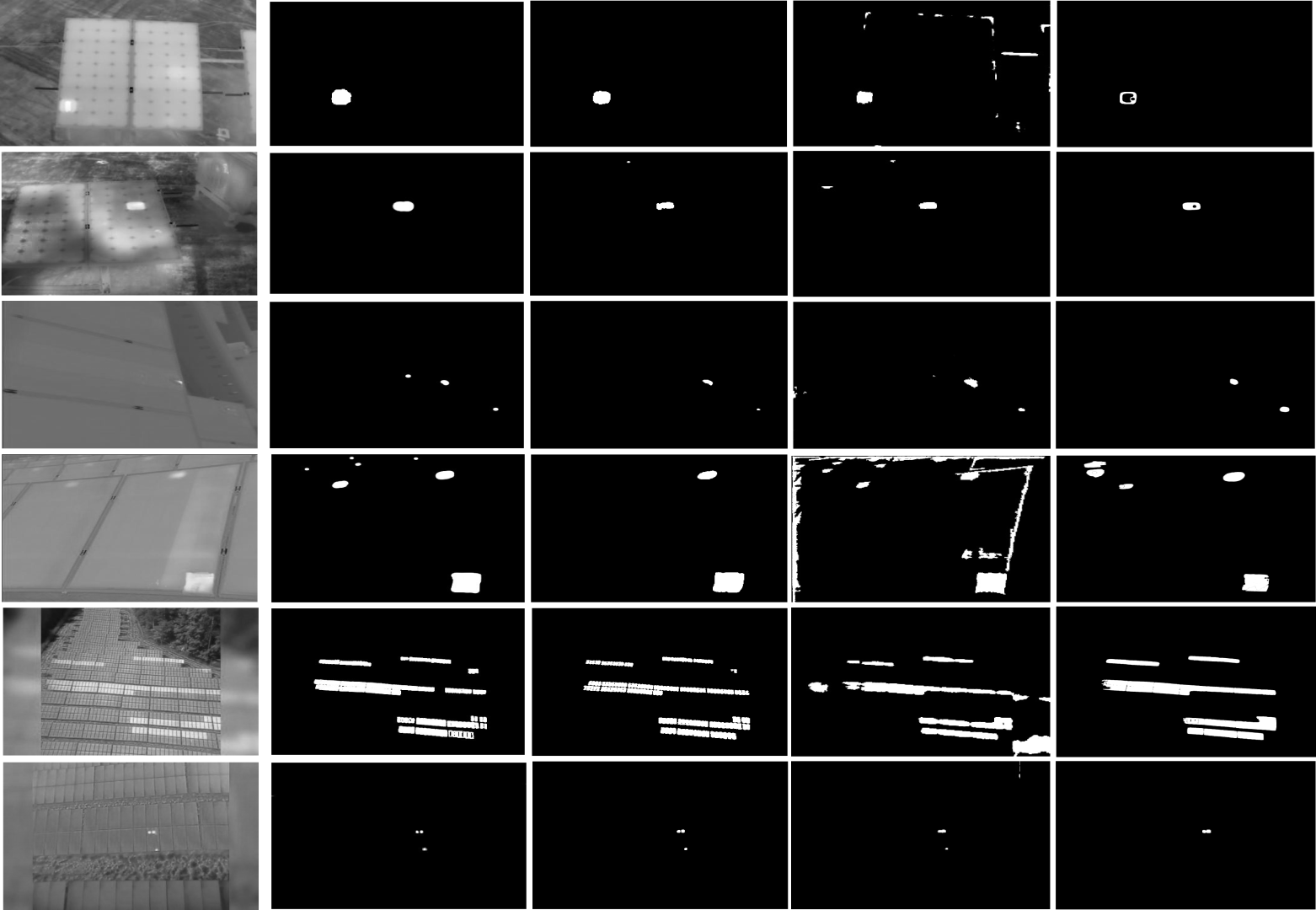}
\caption{Visual segmentation comparison using various methods: Columns 1-5 respectively show the original image, ground truth, proposed, Weka, and Meta segmentation results.}
\label{Figure:066}
\end{figure*}

The segmentation performance of the proposed method, Meta, and Weka approaches was evaluated visually on a set of degraded imagery samples. The evaluation was conducted by comparing the segmented hotspots of the proposed method with those obtained using Meta \cite{kirillov2023segany} and Trainable Weka Segmentation \cite{arganda2017trainable}. The ground truth was used as a reference to compare the segmentation results, as shown in the second column of Figure \ref{Figure:066}.

The segmentation results were compared in terms of visual inspection, and the subjective findings are presented in Figure \ref{Figure:066}. The results indicate that the proposed method outperforms the other two approaches in segmenting abnormality regions. Specifically, the proposed method achieves a more accurate and precise segmentation of the hotspots compared to Meta and Weka.

It is worth noting that the Meta approach performs relatively well but still falls short of the proposed method. Meanwhile, Weka tends to over-segment the images, leading to less accurate segmentation results. Overall, the proposed method offers better visual segmentation results, which can potentially improve the accuracy and efficiency of hotspot detection and characterization tasks.

\subsection{ Evaluation of proposed approach using computer simulations  }

Several statistical methods are widely used to assess the quality of image segmentation. In this paper, we picked the methods including:
Jaccard Index (also known as Intersection over Union or IoU) \cite{benbelkacem2021lung}: This quality measurement calculates the similarity between the segmented region and the ground truth area. The formula for Jaccard Index is given by:

\begin{equation}
\text{Jaccard Index (IoU)} = \frac{\text{TP}}{\text{TP} + \text{FN}}
\end{equation}

Dice Similarity Coefficient: almost similar to Jaccard Index, this coefficient calculates as twice the ratio of the intersection of the segmented and ground truth regions to the sum of their sizes. The formula for Dice Similarity Coefficient is given by:

\begin{equation}
\text{Dice coefficient} = \frac{2 \cdot \text{TP}}{2 \cdot \text{TP} + \text{FP} + \text{FN}}
\end{equation}

Precision and Recall \cite{oulefki2021automatic}: These are commonly used metrics in image segmentation evaluation. Precision measures the accuracy of positive predictions, recall measures the ability to detect positive instances. The formulas for precision and recall are given by:

\begin{equation}
\text{Precision} = \frac{\text{TP}}{\text{TP} + \text{FP}}
\end{equation}

\begin{equation}
\text{Recall (Sensitivity)} = \frac{\text{TP}}{\text{TP} + \text{FN}}
\end{equation}

Rand Index: This method measures the similarity between the segmented region and the ground truth region based on the percentage of agreement in their pixel-wise classifications. The formula for Rand Index is given by :

\begin{equation}
\text{Rand Index} = \frac{\text{TP} + \text{TN}}{\text{TP} + \text{TN} + \text{FP} + \text{FN}}
\end{equation}

\subsection{Performance comparison with existing approaches}

The performance of the proposed segmentation approach was evaluated using metrics such as Jaccard Index (IoU), Dice coefficient, Precision, and Recall measures. The statistical results of the segmentation on a set of imagery with deterioration. In order to compare the performance of the proposed approach efficiently, we also compared the statistical values of the segmented hotspot with those obtained using the proposed, Meta \cite{kirillov2023segany}, and Weka \cite{arganda2017trainable} methods as shown in Table \ref{tab:my-table}.

\begin{table*}[h]
\caption{Statistical segmentation comparison using the proposed against Weka, and Meta segmentation methods}
\label{tab:my-table}
\begin{tabular}{lllllll}
\hline
\textbf{Method~~~~~~~~~~~~~~~~~} &     & \textbf{~~~~~~~~IoU~~~~~~~~} & \textbf{~~~~~~~~DSC~~~~~~~~} & \textbf{~~~~~~~~Prec~~~~~~~~} & \textbf{~~~~~~~~Recall~~~~~~~~} & \textbf{~~~~~~~~RI~~~~~~~~} \\
\hline
\textit{Weka}   & $\mu$  & 0.6869       & 0.5211       & 0.4921        & 0.9660          & 0.6912      \\
                & $\sigma$ & 0.1391       & 0.1943       & 0.2601        & 0.0338          & 0.1324      \\
\hline
\textit{Meta}   & $\mu$  & 0.6212       & 0.6645       & 0.7568        & 0.9791          & 0.6646      \\
                & $\sigma$ & 0.1927       & 0.1433       & 0.2072        & 0.0405          & 0.1512      \\
\hline
\textit{Proposed}   & $\mu$  & 0.7680       & 0.8232       & 0.9045        & 0.9961          & 0.7680      \\
                & $\sigma$ & 0.1457       & 0.0895       & 0.0739        & 0.0056          & 0.1457  \\   
\hline
\end{tabular}
\end{table*}

\begin{figure*}[h]
\centering
\includegraphics[width=12cm]{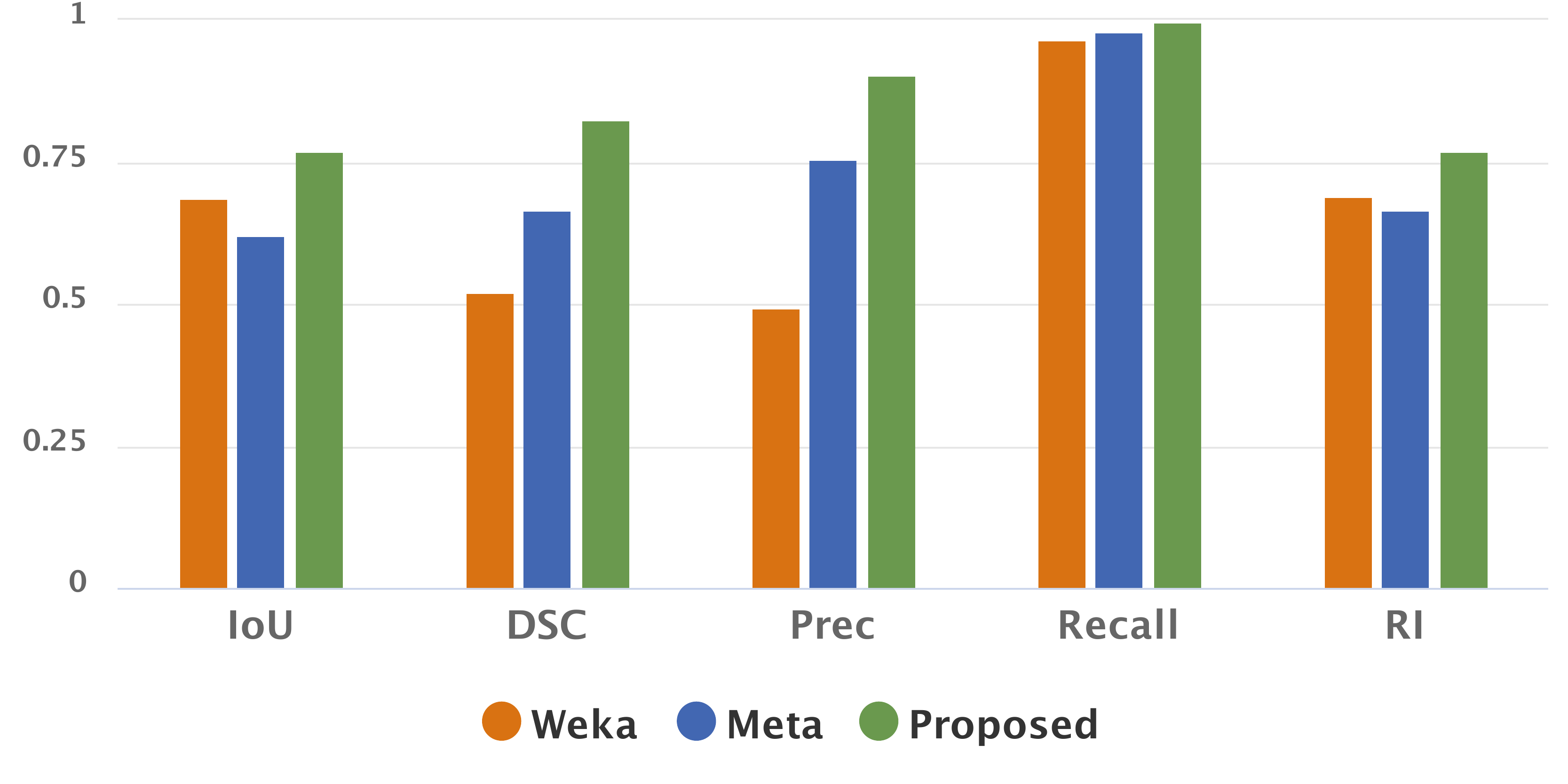}
\caption{Statistical comparison of the proposed segmentation approach with Meta and Weka methods using Jaccard Index (IoU), Dice coefficient, Precision, and Recall metrics.}
\label{Fig:077}
\end{figure*}

Figure \ref{Fig:077} presents the data results of the four matrices observations, discussed earlier, with each method having distinct mean values. As shown in the illustration, the results indicate that the proposed method has the highest mean values, while Meta \cite{kirillov2023segany}, and Weka \cite{arganda2017trainable} methods have slightly lower mean values compared to the proposed method, and Weka has the lowest mean values among the four methods in terms of precision and DSC. These findings suggest that the proposed method likely performs the best among the evaluated methods.

The implementation of the proposed scheme was carried out on a Windows $10$ Pro for Workstations with a $3.7$GHz Intel Core i-9 processor and $32$ GB of RAM, using the latest MATLAB Version.

\subsection{Heat-map chart analysis for determining the visibility of deteriorated areas}

After obtaining the 2D segmentation output, additional processing steps are executed to produce a 3D segmentation output suitable for integration into Blender software. Thus, heat-map chart analysis is a practical procedure picked for determining the visibility of deteriorated areas. It involves creating a color-coded map that highlights the deteriorated areas of the solar panel with high or low-intensity values. The brighter or warmer colors typically indicate areas of higher intensity, while darker or cooler colors represent areas of lower intensity. 

\begin{figure*}[h]
\centering
\includegraphics[width=14cm]{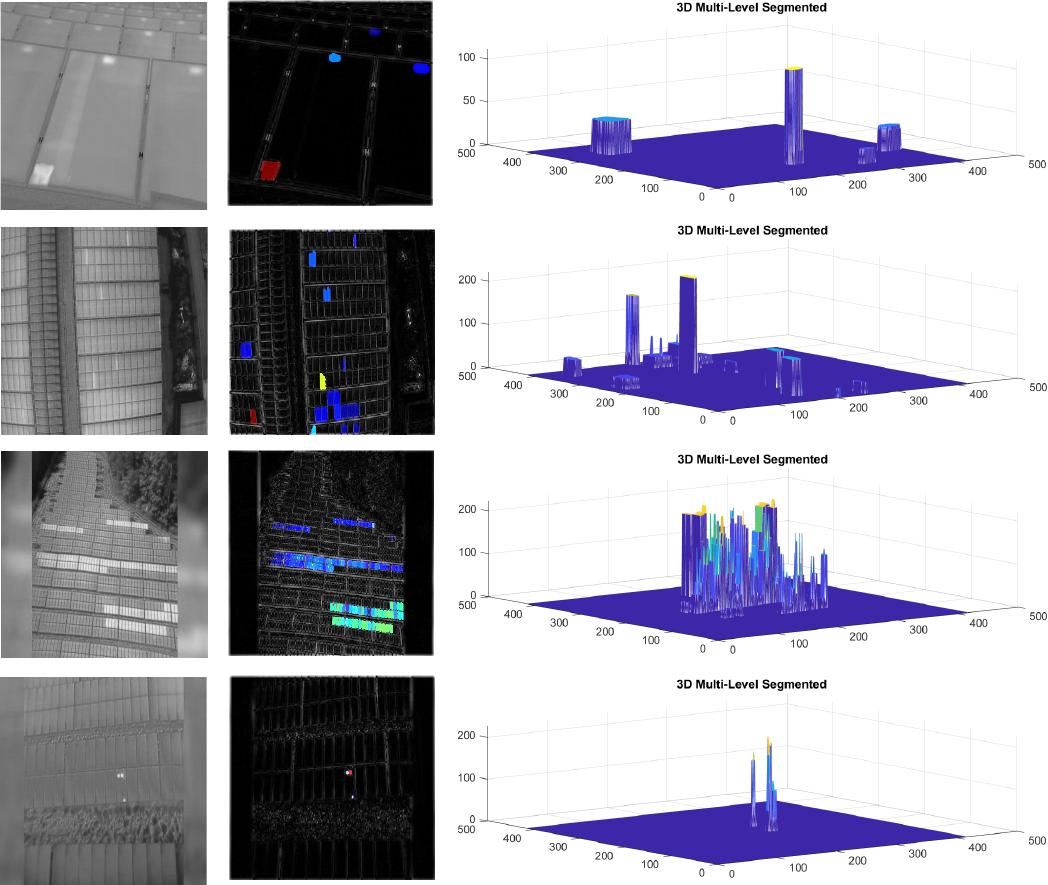}
\caption{Heat-map chart analysis of a deteriorated surface of 4 samples. Black areas indicate intact regions, while colored areas indicate higher levels of surface deterioration.}
\label{Fig:08}
\end{figure*}

For example, Figure \ref{Fig:08} shows a heat-map chart analysis of a deteriorated surface. The darker areas in the image indicate intact regions, while the brighter areas indicate relatively higher levels of deterioration surface.

Overall, heat-map chart analysis is a decisive tool for identifying and visualizing areas of deterioration. This technique can be used in conjunction with other image analysis methods to provide a more comprehensive understanding of the extent and severity of deterioration in a given solar panel.

\subsection{AR visualization Results}

\begin{figure*}[t!]
\centering
\includegraphics[width=0.9\textwidth]{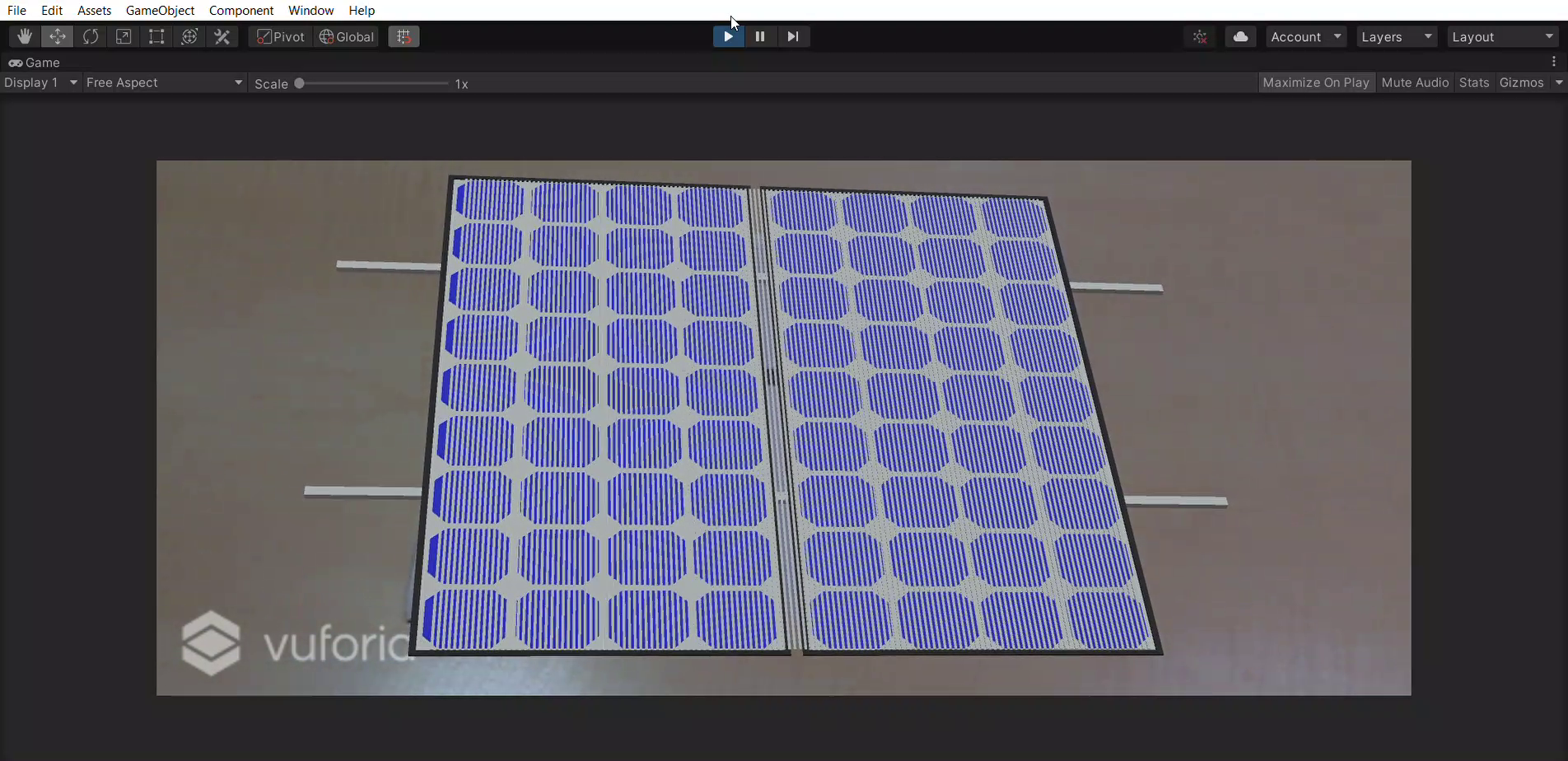}\\
(a)\\
\includegraphics[width=0.9\textwidth]{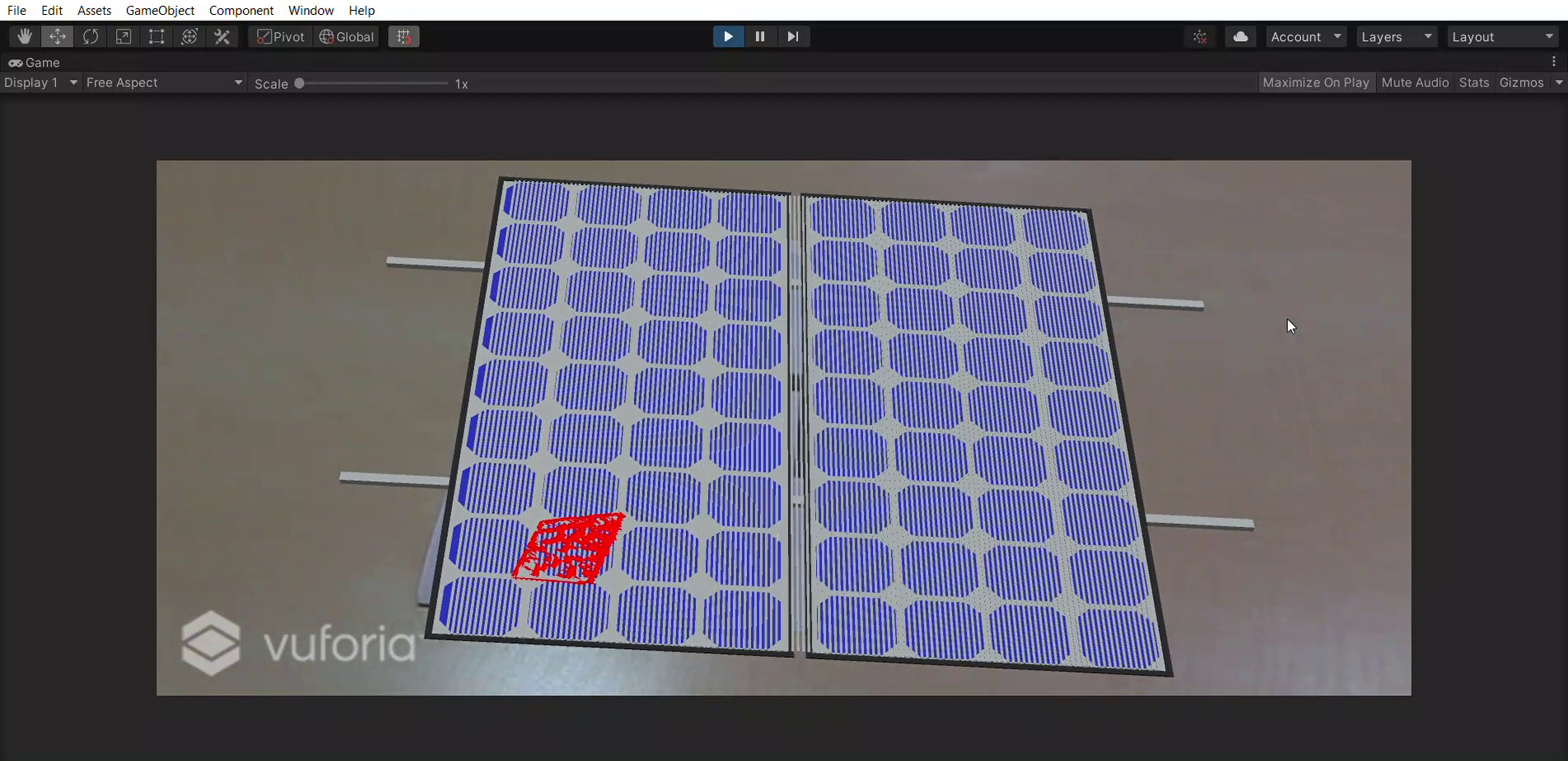}\\
(b)\\
\caption{PV and AR visulations: (a) PV Augmented Reality visualization and (b) AR visualization of the PV deteriorated areas}
\label{PV-AR}
\end{figure*}

Augmented Reality (AR) technology can effectively promote the knowledge and adoption of Photovoltaic (PV) technology and sustainable energy practices by providing an engaging, interactive, and enjoyable experience. The Vuforia SDK for AR rendering was used during the design process, providing a robust and user-friendly tool for analyzing and improving the PV design. The AR rendering of PV visualization is depicted in Figure \ref{PV-AR}(a), and it can be experienced through smartphones, tablets, and AR glasses.

Segmenting the deteriorated areas of PV systems provides significant benefits in terms of accurately locating and diagnosing issues. AR visualization of these segmented areas enables maintenance personnel to view the PV system and its damaged regions in real-time, facilitating precise location and assessment of flaws or damages. Figure \ref{PV-AR}(b) showcases the AR rendering of the visualization of the PV's deteriorated areas. The utilization of AR technology in PV systems has the potential to revolutionize the way we maintain and optimize our renewable energy sources.

\subsection{Discussion}

The interpretation and analysis of the results presented in this study demonstrate the effectiveness of the proposed method for detecting and segmenting deteriorated cells in solar PV panels. Figure \ref{Fig:08} provides valuable information on the thermal performance of PV panels, highlighting the need for segmentation to accurately delineate the affected regions. The analysis reveals that panels with thermal values within the range $T_{min}$=25 $<$ $T_{val}$$<$ $T_{max}$=40 are clean and do not require segmentation, while panels with values outside this range exhibit deteriorated cells.

The segmentation performance of the proposed method was compared with Meta and Weka approaches, and the subjective findings showed that the proposed method outperformed the other two methods in accurately and precisely segmenting abnormality regions. The mean values of the four matrices observations also indicate that the proposed method likely performs the best among the evaluated methods. The heat-map chart analysis proved to be an effective technique for identifying and visualizing areas of deterioration, which can be used in conjunction with other image analysis methods to provide a more comprehensive understanding of the extent and severity of deterioration in a given solar panel.

Furthermore, the study highlights the significance of effective maintenance for solar PV modules. The accurate and efficient detection and segmentation of deteriorated cells can significantly reduce processing time and facilitate the precise location and assessment of flaws or damages, ultimately optimizing the performance of the PV system. Augmented Reality (AR) technology can also play a crucial role in promoting the adoption of Photovoltaic (PV) technology and sustainable energy practices by providing an engaging and interactive experience.

However, this study has limitations that can be addressed in future research. The proposed method was evaluated on a limited set of degraded imagery samples, and more extensive testing is needed to assess its effectiveness on a larger scale. Additionally, the study focused solely on detecting and segmenting deteriorated cells, and future research can explore other aspects of PV maintenance, such as fault detection and diagnosis. Overall, the study offers valuable insights into the importance of effective maintenance for solar PV modules and the potential of advanced technologies like AR in optimizing the performance of renewable energy sources.

\section{Conclusion} \label{Conclusion}
This paper presents a novel approach for detecting abnormalities, such as hot spots and snail trails, in solar photovoltaic (PV) modules using unsupervised sensing algorithms and 3D augmented reality visualization. By facilitating more effective diagnosis and repair procedures, AR can help to lower the cost of PV system maintenance and repair. The proposed segmentation framework and analysis methods are evaluated using computer simulations and real-world image datasets, demonstrating the effectiveness of the approach in identifying dirty areas in solar PV modules. The findings emphasize the importance of regular maintenance to ensure the efficiency and power capacity of solar PV modules. The short-term aim of this work is to detect solar panels in an automatic and real-time manner using drones, which can significantly improve the efficiency of PV module maintenance. The proposed approach could be a game-changer in the field of solar PV maintenance, as it allows for quick and accurate detection of abnormalities without human intervention. This can lead to cost savings, increased energy production, and improved overall performance of solar PV systems. Furthermore, the use of unsupervised sensing algorithms and 3D augmented reality visualization techniques adds a new dimension to the field of solar PV maintenance, opening up possibilities for further research and development in this area.

Future work will focus on exploring the use of generative models (such as ChatGBT and GBT-4) to improve the performance of AR-based visualization and localization of deteriorated areas of solar PV modules built-up in 3D augmented reality holds immense potential for advancing the field of solar PV maintenance. This innovative approach can revolutionize the way we identify, analyze, and address issues in solar PV systems. Typically, by using ChatGBT and integrating advanced machine learning algorithms, the accuracy and efficiency of abnormality detection can significantly be improved. Additionally, by leveraging the vast amount of data collected from solar PV modules, ChatGBT can continuously learn and adapt to different types of deterioration, enabling more precise localization and visualization in 3D augmented reality. Furthermore, the integration of real-time monitoring systems with ChatGBT can enable proactive maintenance strategies by providing timely alerts and notifications when abnormalities are detected. This can help prevent further deterioration and optimize the performance of solar PV systems. Lastly, future research can focus on expanding the applications of ChatGBT in other areas of solar PV system management, such as performance optimization, fault prediction, and energy forecasting. By continuously pushing the boundaries of ChatGBT's capabilities and integrating it into the broader ecosystem of solar PV maintenance, we can unlock new possibilities for improving the sustainability, efficiency, and longevity of solar PV systems.



\begin{thebibliography}{10}
\providecommand{\url}[1]{#1}
\csname url@samestyle\endcsname
\providecommand{\newblock}{\relax}
\providecommand{\bibinfo}[2]{#2}
\providecommand{\BIBentrySTDinterwordspacing}{\spaceskip=0pt\relax}
\providecommand{\BIBentryALTinterwordstretchfactor}{4}
\providecommand{\BIBentryALTinterwordspacing}{\spaceskip=\fontdimen2\font plus
\BIBentryALTinterwordstretchfactor\fontdimen3\font minus
  \fontdimen4\font\relax}
\providecommand{\BIBforeignlanguage}[2]{{%
\expandafter\ifx\csname l@#1\endcsname\relax
\typeout{** WARNING: IEEEtran.bst: No hyphenation pattern has been}%
\typeout{** loaded for the language `#1'. Using the pattern for}%
\typeout{** the default language instead.}%
\else
\language=\csname l@#1\endcsname
\fi
#2}}
\providecommand{\BIBdecl}{\relax}
\BIBdecl

\bibitem{copiaco2023innovative}
A.~Copiaco, Y.~Himeur, A.~Amira, W.~Mansoor, F.~Fadli, S.~Atalla, and S.~S.
  Sohail, ``An innovative deep anomaly detection of building energy consumption
  using energy time-series images,'' \emph{Engineering Applications of
  Artificial Intelligence}, vol. 119, p. 105775, 2023.

\bibitem{himeur2023ai}
Y.~Himeur, M.~Elnour, F.~Fadli, N.~Meskin, I.~Petri, Y.~Rezgui, F.~Bensaali,
  and A.~Amira, ``Ai-big data analytics for building automation and management
  systems: a survey, actual challenges and future perspectives,''
  \emph{Artificial Intelligence Review}, vol.~56, no.~6, pp. 4929--5021, 2023.

\bibitem{alsalemi2022innovative}
A.~Alsalemi, Y.~Himeur, F.~Bensaali, and A.~Amira, ``An innovative edge-based
  internet of energy solution for promoting energy saving in buildings,''
  \emph{Sustainable Cities and Society}, vol.~78, p. 103571, 2022.

\bibitem{elnour2022performance}
M.~Elnour, F.~Fadli, Y.~Himeur, I.~Petri, Y.~Rezgui, N.~Meskin, and A.~M.
  Ahmad, ``Performance and energy optimization of building automation and
  management systems: Towards smart sustainable carbon-neutral sports
  facilities,'' \emph{Renewable and Sustainable Energy Reviews}, vol. 162, p.
  112401, 2022.

\bibitem{himeur2022ai}
Y.~Himeur, M.~Elnour, F.~Fadli, N.~Meskin, I.~Petri, Y.~Rezgui, F.~Bensaali,
  and A.~Amira, ``Ai-big data analytics for building automation and management
  systems: a survey, actual challenges and future perspectives,''
  \emph{Artificial Intelligence Review}, pp. 1--93, 2022.

\bibitem{khan2020investigating}
S.~A.~R. Khan, Z.~Yu, A.~Belhadi, and A.~Mardani, ``Investigating the effects
  of renewable energy on international trade and environmental quality,''
  \emph{Journal of Environmental management}, vol. 272, p. 111089, 2020.

\bibitem{benbelkacem2013augmented}
S.~Benbelkacem, M.~Belhocine, A.~Bellarbi, N.~Zenati-Henda, and M.~Tadjine,
  ``Augmented reality for photovoltaic pumping systems maintenance tasks,''
  \emph{Renewable energy}, vol.~55, pp. 428--437, 2013.

\bibitem{zhang2023evaluating}
T.~Zhang, K.~Nakagawa, and K.~Matsumoto, ``Evaluating solar photovoltaic power
  efficiency based on economic dimensions for 26 countries using a three-stage
  data envelopment analysis,'' \emph{Applied Energy}, vol. 335, p. 120714,
  2023.

\bibitem{rashid2022future}
F.~Rashid and M.~U. Joardder, ``Future options of electricity generation for
  sustainable development: Trends and prospects,'' \emph{Engineering Reports},
  vol.~4, no.~10, p. e12508, 2022.

\bibitem{xia2022mapping}
Z.~Xia, Y.~Li, R.~Chen, D.~Sengupta, X.~Guo, B.~Xiong, and Y.~Niu, ``Mapping
  the rapid development of photovoltaic power stations in northwestern china
  using remote sensing,'' \emph{Energy Reports}, vol.~8, pp. 4117--4127, 2022.

\bibitem{haque2019fault}
A.~Haque, K.~V.~S. Bharath, M.~A. Khan, I.~Khan, and Z.~A. Jaffery, ``Fault
  diagnosis of photovoltaic modules,'' \emph{Energy Science \& Engineering},
  vol.~7, no.~3, pp. 622--644, 2019.

\bibitem{himeur2022next}
Y.~Himeur, M.~Elnour, F.~Fadli, N.~Meskin, I.~Petri, Y.~Rezgui, F.~Bensaali,
  and A.~Amira, ``Next-generation energy systems for sustainable smart cities:
  Roles of transfer learning,'' \emph{Sustainable Cities and Society}, p.
  104059, 2022.

\bibitem{dwivedi2020advanced}
P.~Dwivedi, K.~Sudhakar, A.~Soni, E.~Solomin, and I.~Kirpichnikova, ``Advanced
  cooling techniques of pv modules: A state of art,'' \emph{Case studies in
  thermal engineering}, vol.~21, p. 100674, 2020.

\bibitem{ibne2023impact}
F.~ibne Mahmood and G.~TamizhMani, ``Impact of different backsheets and
  encapsulant types on potential induced degradation (pid) of silicon pv
  modules,'' \emph{Solar Energy}, vol. 252, pp. 20--28, 2023.

\bibitem{arosh2023composite}
S.~Arosh, K.~Ghosh, D.~K. Dheer, and S.~Prakash, ``Composite imagery-based
  non-uniform illumination sensing for system health monitoring of solar power
  plants,'' \emph{Journal of Solar Energy Engineering}, vol. 145, no.~1, p.
  011009, 2023.

\bibitem{PALMER2019989}
\BIBentryALTinterwordspacing
J.~Palmer, ``Smog casts a shadow on solar power,'' \emph{Engineering}, vol.~5,
  no.~6, pp. 989--990, 2019. [Online]. Available:
  \url{https://www.sciencedirect.com/science/article/pii/S2095809919308598}
\BIBentrySTDinterwordspacing

\bibitem{dhimish201970}
M.~Dhimish, ``70\% decrease of hot-spotted photovoltaic modules output power
  loss using novel mppt algorithm,'' \emph{IEEE Transactions on Circuits and
  Systems II: Express Briefs}, vol.~66, no.~12, pp. 2027--2031, 2019.

\bibitem{aghaei2022review}
M.~Aghaei, A.~Fairbrother, A.~Gok, S.~Ahmad, S.~Kazim, K.~Lobato, G.~Oreski,
  A.~Reinders, J.~Schmitz, M.~Theelen \emph{et~al.}, ``Review of degradation
  and failure phenomena in photovoltaic modules,'' \emph{Renewable and
  Sustainable Energy Reviews}, vol. 159, p. 112160, 2022.

\bibitem{trongtirakul2022unsupervised}
T.~Trongtirakul and S.~Agaian, ``Unsupervised and optimized thermal image
  quality enhancement and visual surveillance applications,'' \emph{Signal
  Processing: Image Communication}, vol. 105, p. 116714, 2022.

\bibitem{himeur2021artificial}
Y.~Himeur, K.~Ghanem, A.~Alsalemi, F.~Bensaali, and A.~Amira, ``Artificial
  intelligence based anomaly detection of energy consumption in buildings: A
  review, current trends and new perspectives,'' \emph{Applied Energy}, vol.
  287, p. 116601, 2021.

\bibitem{himeur2020novel}
Y.~Himeur, A.~Alsalemi, F.~Bensaali, and A.~Amira, ``A novel approach for
  detecting anomalous energy consumption based on micro-moments and deep neural
  networks,'' \emph{Cognitive Computation}, vol.~12, pp. 1381--1401, 2020.

\bibitem{fahimipirehgalin2021automatic}
M.~Fahimipirehgalin, E.~Trunzer, M.~Odenweller, and B.~Vogel-Heuser,
  ``Automatic visual leakage detection and localization from pipelines in
  chemical process plants using machine vision techniques,''
  \emph{Engineering}, vol.~7, no.~6, pp. 758--776, 2021.

\bibitem{henry2020automatic}
C.~Henry, S.~Poudel, S.-W. Lee, and H.~Jeong, ``Automatic detection system of
  deteriorated pv modules using drone with thermal camera,'' \emph{Applied
  Sciences}, vol.~10, no.~11, p. 3802, 2020.

\bibitem{masita202275mw}
K.~Masita, A.~Hasan, and T.~Shongwe, ``75mw ac pv module field anomaly
  detection using drone-based ir orthogonal images with res-cnn3 detector,''
  \emph{IEEE Access}, vol.~10, pp. 83\,711--83\,722, 2022.

\bibitem{al2022interactive}
A.~Al-Kababji, A.~Alsalemi, Y.~Himeur, R.~Fernandez, F.~Bensaali, A.~Amira, and
  N.~Fetais, ``Interactive visual study for residential energy consumption
  data,'' \emph{Journal of Cleaner Production}, vol. 366, p. 132841, 2022.

\bibitem{AR-main}
\BIBentryALTinterwordspacing
D.~Mourtzis, V.~Siatras, and J.~Angelopoulos, ``Real-time remote maintenance
  support based on augmented reality (ar),'' \emph{Applied Sciences}, vol.~10,
  no.~5, 2020. [Online]. Available:
  \url{https://www.mdpi.com/2076-3417/10/5/1855}
\BIBentrySTDinterwordspacing

\bibitem{al2021solar}
O.~A. Al-Shahri, F.~B. Ismail, M.~Hannan, M.~H. Lipu, A.~Q. Al-Shetwi,
  R.~Begum, N.~F. Al-Muhsen, and E.~Soujeri, ``Solar photovoltaic energy
  optimization methods, challenges and issues: A comprehensive review,''
  \emph{Journal of Cleaner Production}, vol. 284, p. 125465, 2021.

\bibitem{al2020energy}
A.~Al-Kababji, A.~Alsalemi, Y.~Himeur, F.~Bensaali, A.~Amira, R.~Fernandez, and
  N.~Fetais, ``Energy data visualizations on smartphones for triggering
  behavioral change: Novel vs. conventional,'' in \emph{2020 2nd Global Power,
  Energy and Communication Conference (GPECOM)}.\hskip 1em plus 0.5em minus
  0.4em\relax IEEE, 2020, pp. 312--317.

\bibitem{benbelkacem2022covi3d}
S.~Benbelkacem, A.~Oulefki, S.~Agaian, N.~Zenati-Henda, T.~Trongtirakul,
  D.~Aouam, M.~Masmoudi, and M.~Zemmouri, ``Covi3d: Automatic covid-19 ct
  image-based classification and visualization platform utilizing virtual and
  augmented reality technologies,'' \emph{Diagnostics}, vol.~12, no.~3, p. 649,
  2022.

\bibitem{AR-PV-Main}
N.~Zenati, M.~Hamidia, A.~Bellarbi, and S.~Benbelkacem, ``E-maintenance for
  photovoltaic power system in algeria,'' in \emph{2015 IEEE International
  Conference on Industrial Technology (ICIT)}, 2015, pp. 2594--2599.

\bibitem{alsafasfeh2018unsupervised}
M.~Alsafasfeh, I.~Abdel-Qader, B.~Bazuin, Q.~Alsafasfeh, and W.~Su,
  ``Unsupervised fault detection and analysis for large photovoltaic systems
  using drones and machine vision,'' \emph{Energies}, vol.~11, no.~9, p. 2252,
  2018.

\bibitem{shihavuddin2021image}
A.~Shihavuddin, M.~R.~A. Rashid, M.~H. Maruf, M.~A. Hasan, M.~A. ul~Haq, R.~H.
  Ashique, and A.~Al~Mansur, ``Image based surface damage detection of
  renewable energy installations using a unified deep learning approach,''
  \emph{Energy Reports}, vol.~7, pp. 4566--4576, 2021.

\bibitem{zyout2020detection}
I.~Zyout and A.~Oatawneh, ``Detection of pv solar panel surface defects using
  transfer learning of the deep convolutional neural networks,'' in \emph{2020
  Advances in Science and Engineering Technology International Conferences
  (ASET)}.\hskip 1em plus 0.5em minus 0.4em\relax IEEE, 2020, pp. 1--4.

\bibitem{abuqaaud2020novel}
K.~A. Abuqaaud and A.~Ferrah, ``A novel technique for detecting and monitoring
  dust and soil on solar photovoltaic panel,'' in \emph{2020 Advances in
  Science and Engineering Technology International Conferences (ASET)}.\hskip
  1em plus 0.5em minus 0.4em\relax IEEE, 2020, pp. 1--6.

\bibitem{pierdicca2020automatic}
R.~Pierdicca, M.~Paolanti, A.~Felicetti, F.~Piccinini, and P.~Zingaretti,
  ``Automatic faults detection of photovoltaic farms: solair, a deep
  learning-based system for thermal images,'' \emph{Energies}, vol.~13, no.~24,
  p. 6496, 2020.

\bibitem{alfaro2019dataset}
E.~Alfaro-Mejia, H.~Loaiza-Correa, E.~Franco-Mejia, A.~D. Restrepo-Giron, and
  S.~E. Nope-Rodriguez, ``Dataset for recognition of snail trails and hot spot
  failures in monocrystalline si solar panels,'' \emph{Data in brief}, vol.~26,
  p. 104441, 2019.

\bibitem{gallardo2017aerial}
S.~Gallardo-Saavedra, E.~Franco-Mejia, L.~Hern{\'a}ndez-Callejo,
  {\'O}.~Duque-P{\'e}rez, H.~Loaiza-Correa, and E.~Alfaro-Mejia, ``Aerial
  thermographic inspection of photovoltaic plants: analysis and selection of
  the equipment,'' in \emph{Proceedings of the 2017 Proceedings ISES Solar
  World Congress, IEA SHC, Abu Dhabi, UAE}, vol.~29, 2017.

\bibitem{solar-panel-infrared-images-v5}
Roboflow, ``Solar panel infrared images dataset,''
  \url{https://universe.roboflow.com/dataset/solar-panel-infrared-images-v5},
  February 6 2023, exported via Roboflow on July 27, 2022.

\bibitem{SPII_DatasetName}
\BIBentryALTinterwordspacing
RoboFlow, ``solar-panel-infrared-images - v5 2022-07-27 7:03pm,'' RoboFlow,
  2023, exported from roboflow.com on February 6, 2023 at 8:54 AM GMT.
  [Online]. Available: \url{https://universe.roboflow.com}
\BIBentrySTDinterwordspacing

\bibitem{vega2020solar}
J.~J. Vega~D{\'\i}az, M.~Vlaminck, D.~Lefkaditis, S.~A. Orjuela~Vargas, and
  H.~Luong, ``Solar panel detection within complex backgrounds using thermal
  images acquired by uavs,'' \emph{Sensors}, vol.~20, no.~21, p. 6219, 2020.

\bibitem{SolidWorks}
D.~Systèmes, ``Solidworks,'' \url{https://www.solidworks.com/fr}, Accessed
  2023.

\bibitem{Blender}
B.~Foundation, ``Blender,'' \url{https://www.blender.org/}, Accessed 2023.

\bibitem{Vuforia}
P.~Inc., ``Vuforia,''
  \url{https://www.ptc.com/en/products/augmented-reality/vuforia}, Accessed
  2023.

\bibitem{Unity}
U.~Technologies, ``Unity,'' \url{https://www.unity.com/}, Accessed 2023.

\bibitem{kirillov2023segany}
A.~Kirillov, E.~Mintun, N.~Ravi, H.~Mao, C.~Rolland, L.~Gustafson, T.~Xiao,
  S.~Whitehead, A.~C. Berg, W.-Y. Lo, P.~Doll{\'a}r, and R.~Girshick, ``Segment
  anything,'' \emph{arXiv:2304.02643}, 2023.

\bibitem{arganda2017trainable}
I.~Arganda-Carreras, V.~Kaynig, C.~Rueden, K.~W. Eliceiri, J.~Schindelin,
  A.~Cardona, and H.~Sebastian~Seung, ``Trainable weka segmentation: a machine
  learning tool for microscopy pixel classification,'' \emph{Bioinformatics},
  vol.~33, no.~15, pp. 2424--2426, 2017.

\bibitem{benbelkacem2021lung}
S.~Benbelkacem, A.~Oulefki, S.~Agaian, T.~Trongtirakul, D.~Aouam,
  N.~Zenati-Henda, and K.~Amara, ``Lung infection region quantification,
  recognition, and virtual reality rendering of ct scan of covid-19,'' in
  \emph{Multimodal Image Exploitation and Learning 2021}, vol. 11734.\hskip 1em
  plus 0.5em minus 0.4em\relax SPIE, 2021, pp. 123--132.

\bibitem{oulefki2021automatic}
A.~Oulefki, S.~Agaian, T.~Trongtirakul, and A.~K. Laouar, ``Automatic covid-19
  lung infected region segmentation and measurement using ct-scans images,''
  \emph{Pattern recognition}, vol. 114, p. 107747, 2021.

\end{thebibliography}


\end{document}